\documentclass{article} 
\usepackage{iclr2025_conference,times}

\usepackage{microtype}
\usepackage{graphicx}
\usepackage{subfigure}
\usepackage{booktabs} 

\usepackage{hyperref}

\usepackage{amsmath}
\usepackage{amssymb}
\usepackage{mathtools}
\usepackage{amsthm}

\usepackage[capitalize,noabbrev]{cleveref}

\theoremstyle{plain}

\theoremstyle{definition}

\theoremstyle{remark}

\usepackage[textsize=tiny]{todonotes}

\usepackage{multirow} 
\usepackage{color, soul}
\sethlcolor{yellow}
\setstcolor{red}
\setulcolor{green}

\definecolor{Magenta}{rgb}{0.8, 0.1, 0.6}

\usepackage{amsmath}
\usepackage{amsfonts}
\usepackage{amssymb}
\usepackage{wrapfig}
\usepackage{subcaption}
\usepackage{multirow}
 \usepackage{mathtools} 

\usepackage{verbatim}

\usepackage{anyfontsize}

\usepackage{microtype}
\usepackage{graphicx}
\usepackage{booktabs} 
\usepackage{multirow}
\usepackage{amsmath,amssymb}
\usepackage{booktabs}
\usepackage{caption,subcaption}

\usepackage{xcolor}
\definecolor{mygreen}{HTML}{3cb44b}
\definecolor{skyblue}{HTML}{beffff}
\definecolor{lightgreen}{HTML}{90ee90}

\usepackage{color, colortbl}

\definecolor{emerald}{rgb}{0.31, 0.78, 0.37}

\usepackage{tcolorbox}
\usepackage{enumitem}
\setitemize{itemsep=10pt,topsep=0pt,parsep=0pt,partopsep=0pt}
\usepackage{colortbl}

\usepackage{xcolor}
\definecolor{mygreen}{HTML}{3cb44b}
\colorlet{myyellow}{green!10!orange!90!}
\makeatletter

\usepackage{tikz}
\usetikzlibrary{arrows,shapes,snakes,automata,backgrounds,fit,petri}
\usepackage{adjustbox}

\newcommand{\RN}[1]{%
	\textup{\lowercase\expandafter{\it \romannumeral#1}}%
}
\usepackage{tabu}

\newcommand{\beq}{\vspace{0mm}\begin{equation}}
\newcommand{\eeq}{\vspace{0mm}\end{equation}}
\newcommand{\beqs}{\vspace{0mm}\begin{eqnarray}}
\newcommand{\eeqs}{\vspace{0mm}\end{eqnarray}}
\newcommand{\barr}{\begin{array}}
\newcommand{\earr}{\end{array}}

\usepackage{color, colortbl}
\definecolor{Gray}{gray}{0.93}

\usepackage{lipsum}

\usepackage{pifont}

\usepackage{makecell}

\usepackage{xcolor,amsmath}
\usepackage[linesnumbered,ruled,vlined]{algorithm2e}
\DontPrintSemicolon

\usepackage{xcolor}
\definecolor{mygreen}{HTML}{3cb44b}

\SetKwComment{Comment}{\color{green!50!black}\# }{}

\SetKwProg{Function}{def}{:}{}

\SetKwProg{For}{for}{:}{}
\SetKwProg{If}{if}{:}{}

\usepackage{cleveref}
\usepackage{alltt}
\tcbuselibrary{most}
\tcbset{
  aibox/.style={
    width=\textwidth,
    top=10pt,
    colback=white,
    colframe=black,
    colbacktitle=black,
    enhanced,
    center,
    attach boxed title to top left={yshift=-0.1in,xshift=0.15in},
    boxed title style={boxrule=0pt,colframe=white,},
  }
}
\newtcolorbox{AIbox}[2][]{aibox,title=#2,#1}
\newlength\savewidth

\title{Data Metabolism: An Efficient Data Design Scheme for Vision Language Models}

\author{Jingyuan Zhang, \
Hongzhi Zhang, \
Zhou Haonan, \
Chenxi Sun, \
Xingguang Ji, \\
\textbf{Jiakang Wang}, \
\textbf{Fanheng Kong}, \ 
\textbf{Yahui Liu}, \
\textbf{Qi Wang}, \
\textbf{Fuzheng Zhang}
\\ 
Kuaishou Technology
}

\newcommand{\capybara}{\textit{Capybara-VL}}
\newcommand{\model}{\textit{Capybara-VL-7B}}
\newcommand{\dm}{\textit{Data Metabolism}}

\iclrfinalcopy 

\begin{document}
\maketitle

\begin{abstract}
Data curation plays a crucial role in training powerful Visual Language Models (VLMs). 
In this work, we introduce the concept of \dm \ and present our data-centric framework to build VLMs throughout the development lifecycle.
Starting from a standard model architecture, we discuss and provide insights into two crucial development steps: data curation and iteration, forming a closed-loop system that continuously improves model performance.
We show a detailed codebook on how to process existing massive datasets and build user-specific data flywheel.
As a demonstration, we release a VLM, named \model\footnote{We name our model \capybara\  and provide support to anyone who needs it. The model and code are available at \url{https://github.com/zjy-ucas/Capybara-VL}}, which excels in typical multimodal tasks (\textit{e.g.}, visual question answering, scientific reasoning and text-rich tasks). 
Despite its relatively compact size, \model\ surpasses several open-source models that are up to $\times$10 times larger in size. Moreover, it achieves results that are on par with those of several leading proprietary models, demonstrating its remarkable competitiveness.
These results highlight the power of our data-centric framework and the potential of training smaller and more efficient VLMs.
\end{abstract}

\section{Introduction}
Visual Language Models (VLMs) represent a pivotal advancement in Artificial Intelligence (AI), integrating visual perception with language understanding~\citep{gpt4,anthropic2024claude35,li2023blip,gao2024sphinx,lu2024deepseek}. 
Owing to the notable contributions from both commercial models~\citep{gpt4,geminiteam2024gemini15unlockingmultimodal} 
and open-source alternatives with high performance~\citep{chen2023internvl,wang2024qwen2vlenhancingvisionlanguagemodels,llama3.2v},
VLMs have emerged with numerous powerful and interesting behaviors and applications, including visual question answering~\citep{antol2015vqa,mathew2021docvqa}, video understanding~\citep{videollava,Maaz2023VideoChatGPT,zhao2025llava,zhang2025llava}, multimodal reasoning~\citep{lu2023mathvista,zhang2024mathverse}, and transformative potential in industrial applications~\citep{chen2024vlmsplayactionroleplaying,long2024vlmmpcvisionlanguagefoundation,chen2024huatuogpt}.

As we all know, both the training pipeline and corresponding datasets play a crucial role in the final performance of the VLMs.  
In recent research work~\citep{hu2024mplugdocowl15unifiedstructure,shang2024pixelstokensrevisitingobject,li2024tokenpackerefficientvisualprojector,tong2024cambrian,zhao2025llava,li2025llava}, when releasing a model, we observe that researchers always release a corresponding meticulously crafted dataset at the same time. 
However, with only those open-source models or public data mixtures, it is not clear how to design the data process in an efficient manner, leading to an important yet unsolved problem. 
Especially, once the data is selected, it will not change anymore during the training process, thus the final model is very likely to be affected by the problematic data. 
Once we want to eliminate the problematic data and update the training, the process is cumbersome and the cost is also pretty high.

\begin{figure*}[h]
    \centering
    \includegraphics[width=\textwidth]{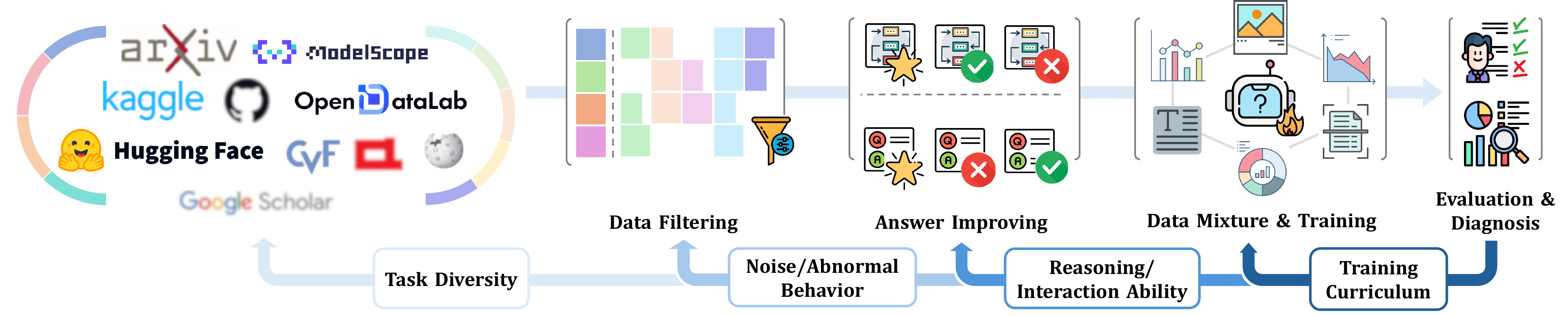}
    \caption{A concise illustration of data metabolism, where training data are iteratively collected, filtered, and improved, while diagnosing the model to guide actions for the next iteration.
    }
    \label{fig:framework_tiny}
\end{figure*}

In this work, we claim that both data processing and diagnosing are essential in the data design of VLMs. Thereby, we provide a detailed data processing recipe for researchers
and demonstrate how to diagnose the effectiveness of the training data according to various model behaviors. 
We name this multimodal data processing and diagonosing as \textbf{\textit{Data Metabolism}}, which is inspired by the biological concept of \textit{metabolism}~\citep{fell1997understanding,wang2009metabolism}.
In biological systems, metabolism is the set of life-sustaining chemical processes that continuously occur and allow life normal functioning. 
It generally includes processes that break down nutrients from food for energy (\textit{i.e.}, the \textit{Anabolism Phase}), and processes that build and repair the body by consuming the energy (\textit{i.e.}, the \textit{Catabolism Phase}).
Here, we borrow the concept in a way that the \textit{\textbf{Data Anabolism Phase}} means how to construct high-quality data that are nutrients for model training, while the \textit{\textbf{Data Catabolism Phase}} means how to diagnostic the model when it demonstrates different issues. 
As shown in Figure~\ref{fig:framework_tiny},
the distinct feature of our proposed method is that the data and models can be dynamically and reasonably adjusted during the optimizing procedures.

Using these ideas, we show how to train a VLM and validate and improve its performance with our introduced data metabolism process. We employ a standard model architecture and training stages. 
Through extensive examinations of various multimodal image understanding benchmarks, we iteratively refine our training data, and finally reach a VLM with a performance comparable to several proprietary models and surpasses open-source models with significantly larger parameter counts.
Taking into account the effectiveness, we hope that our piloting work will lead to more explorations in the field of data design schemes for VLMs.

Our contribution could be summarized as follows:  
\begin{enumerate}  
\setlength\itemsep{0.1pt}
    \item We introduce the concept of \textbf{\dm}, as a cookbook on how to process existing massive datasets, and the principle for building strong VLMs.  
    \item We detail each stage of the \dm \ development lifecycle, 
    from initial creation to iterative refinement, offering actionable insights for data-driven model improvement.  
    \item We release \model \ that achieves competitive performance on multiple multimodal tasks, including visual question answering, scientific reasoning, and text-rich tasks.  
\end{enumerate}  

\section{Preliminaries}

To verify the effectiveness of our proposed \textit{Data Metabolism}, we build upon a standard VLM architecture~\citep{liu2023llava} consisting of three main components: a SigLIP visual encoder~\citep{yao2024minicpmvgpt4vlevelmllm}, a Qwen2.5-instruct-7b language model, and an MLP with 2x2 pooling as the modality adaptor. 
Following~\citep{xu2024llavauhdlmmperceivingaspect}, we apply \textit{any resolution} (AnyRes) method and interpolate the position embeddings of the ViT by following NaViT~\citep{dehghani2023patchnpacknavit} to process images with arbitrary resolution and aspect ratios.

Before introducing the roadmap to utilize the \textit{Data Metabolism} into the training process, we briefly summarize the training of a VLM into three stages:

\textbf{Stage 1: Visual Concept Alignment} Previous work has revealed that it is crucial to establish a connection between the representation space of images and the LLM textual embedding space. Therefore, we utilize the projection approach~\citep{bai2023qwen,liu2023llava,liu2024improved} to construct such a connection.  
In the first stage, we train an adaptor only to align high-level concepts between the vision encoder and the LLM. The rest parameters remain frozen and AnyRes is not applied in this stage. 

\textbf{Stage 2: Detailed Image Understanding} 
The second stage improves the ability to understand fine-grained visual concepts and handle high-resolution images. Following to the AnyRes method, images are split with up to four sub-images, allowing the model to handle resolutions up to 896×896 pixels with arbitrary aspect ratios. 
Generally, the training data of this stage includes 
high-resolution images with dense captions and text-rich content such as documents, charts, and tables, along with core visual knowledge as the training dataset.
All model parameters are unfrozen during this stage.

\textbf{Stage 3: Instruction Tuning}  
The final stage aims to enhance multimodal interaction and reasoning abilities, as well as other general tasks. In this stage, images are divided into up to nine sub-images, supporting resolutions of up to 1344×1344 pixels with arbitrary aspect ratios. 
The training data at this stage includes diverse datasets (\textit{e.g.}, instruction following, general QA, dialogue, and reasoning tasks). All parameters are trainable at this stage.

\section{Data Metabolism}
\label{sec:data_metabolism}

\textit{Metabolism} is a biological concept that reveals the set of life-sustaining chemical reactions in organisms. Interestingly, we first introduce this concept into the field of optimizing VLMs. We highlight that there are two crucial phases in our proposed optimizations:

\textbf{Data Anabolism Phase} 
It refers to procedures that transform raw data into a suitable state for training, including:
1) \textit{Filtering processing}: Reducing duplicates, removing low-relevance image-text pairs, and filtering low-quality responses (see details in Section~\ref{sec:data_filtering} and Appendix~\ref{appendix:data_filtering}).
2) \textit{Data quality improvement}: Enhancing data through prompt isolation to reduce format interference, rewriting answers for verbosity and alignment with human preferences, and generating Chain-of-Thought (CoT) answers to boost multimodal reasoning (see details in  Section~\ref{sec:data_improvement} and Appendix~\ref{appendix:data_improvement}).

\textbf{Data Catabolism Phase} 
It refers to the actions taken after analyzing performance feedback from training a VLM with data from the anabolism phase. This involves two key steps: (1) Diagnosis, using protocol tools to identify unexpected or suboptimal behaviors (see Section~\ref{subsec:diagnosis}); and (2) Updating, where we refine the dataset by adding effective new data and removing redundant samples based on the diagnosis feedback (see Section~\ref{subsec:updating_data}).

\begin{figure*}[h]
    \centering
    \includegraphics[width=\textwidth]{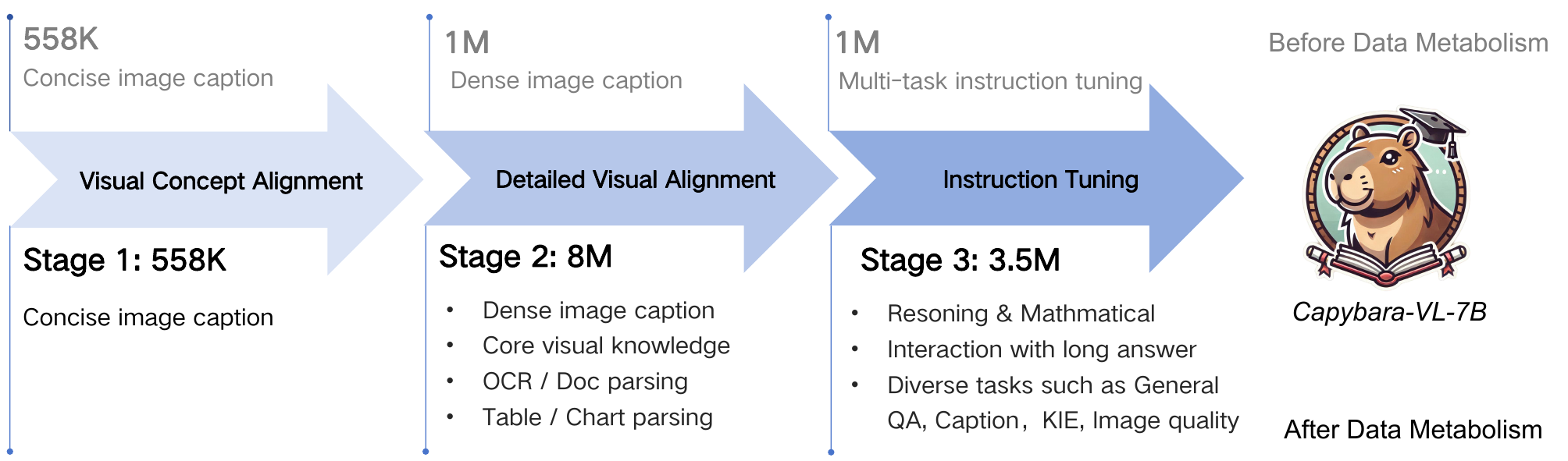}
    \caption{Comparisons on the data used in the three training stages before and after \textit{Data Metabolism}. 
    We refer to Appendix for the details on the data categories (\textit{i.e.}, Table~\ref{tab:data_mixture_stage2} and Table~\ref{tab:data_mixture_stage3}).
    }
    \label{fig:data_stage}
\end{figure*}

We first train a model using 2.5M open-source data as an initial version.
For Stage 1, we use 558k coarse-grained caption data~\citep{liu2023llava}, for Stage 2, 1M dense caption data~\citep{chen2024allava}, and for Stage 3, 1M instruction data~\citep{chen2023sharegpt4v}. Building on this foundation, we diagnose model issues and iterate on the data. Since Stage 1 focuses on basic visual concept alignment, our main iterations are focused on the data for Stage 2 and Stage 3.
The data mixture of the initial and final models is summarized in Fiugre~\ref{fig:data_stage}, with the final mixtures for Stage 2 and Stage 3 detailed in Tables~\ref{tab:data_mixture_stage2} and~\ref{tab:data_mixture_stage3} in Appendix, respectively.

Over the course of iterations, we made four major modifications: (1) add 1M samples for stage 3 from diverse sources, enhancing performance across various dimensions; (2) In Stage 2, we add a large amount of OCR, document, table, and chart samples, improving the perception ability in text-rich scenarios. We then incorporate diverse text-rich understanding and reasoning data in stage 3, enhancing the model’s ability to process text-rich images.; (3) include 1M samples of CoT reasoning and answer rephrasing data, boosting the model's reasoning and interaction capabilities; and (4) strict quality assessment with GPT-4o, enabling better filtering of low-quality data and leading to overall performance improvements across multiple tasks.

\section{Data Anabolism Phase}\label{anabolism}
This section describes the data preparation process, including data collecting, filtering, and improving response quality by rewriting answers of existing samples. 

We collect a large amount of open-source data, spanning datasets that focus on a specific task and data collections for prior open-source works~\citep{laurenccon2024matters,lu2024ovisstructuralembeddingalignment,tong2024cambrian,li2024llavaonevisioneasyvisualtask}. In total, we gather more than 38M samples from more than 200 distinct sources. After applying deduplication and filtering, we generate additional samples by rewriting answers in the training data. This results in a final dataset of approximately 12M high-quality samples.

The collected data is organized by task type, question type, and response source. As shown in Table~\ref{tab:data_mixture_stage3}, we categorize these data into ten major task types. For question types, we classify the data according to the nature of their answer format: yes/no, multiple choice, short-form answers (phrases), long-form answers (free-form), and captions. Response sources are grouped into ground truth, human annotations, and model-generated outputs. For model-generated responses, we further label the specific model or pipeline used for generation. These classifications facilitate dataset review, data filtering, and answer refinement.

\paragraph{Data Filterling}
\label{sec:data_filtering}
When applying data filtering, we focus on three aspects: data diversity, the relevance between images and questions, and the quality of responses. Considering the overlap between open-source data, we first remove duplicate training samples. If both the image and the question are the same, we only keep one sample.
We found that some training data contains questions that do not relate to the images or even conflict with them. To address this, we use the Qwen2-VL series model for scoring. For samples that are not related to the image but contain scientific knowledge, we discard the image and treat the sample as text-only data. We use the same method for questions that can be answered without visual input. We discard other unrelated or conflicting samples.
In the data quality screening stage, we first use heuristic rules to filter out samples with unexpected behavior or repeated content in the responses. For free-form questions, we use GPT4o to score the response quality, filtering out low-quality synthetic data that are overlooked. See more details in Appendix~\ref{appendix:data_filtering}.

\paragraph{Answer Improving}
\label{sec:data_improvement}
In addition to selecting high-quality training data, we enhance the model's capabilities in interaction and reasoning tasks by rewriting answers of existing questions to provide richer supervision signals. Since the training data includes a variety of multi-task datasets, which benefit overall model performance, the fixed answer formats often result in short responses. To address this, we use prompt isolation to distinguish task types, allowing for task adaptability while minimizing negative impacts on interaction and reasoning tasks. Specifically, we sample 10\% of the original instruction-tuning data, generating extended responses based on short answers and filtering for quality. For reasoning-related tasks, we create step-by-step chain-of-thought (CoT) training data and apply rule-based and quality model filtering to obtain high-quality responses, further improving the model's reasoning ability. More details are shown in Appendix~\ref{appendix:data_improvement}.
\section{Data Catabolism Phase}\label{catabolism}
\begin{figure*}[h]
    \centering
    \includegraphics[width=\textwidth]{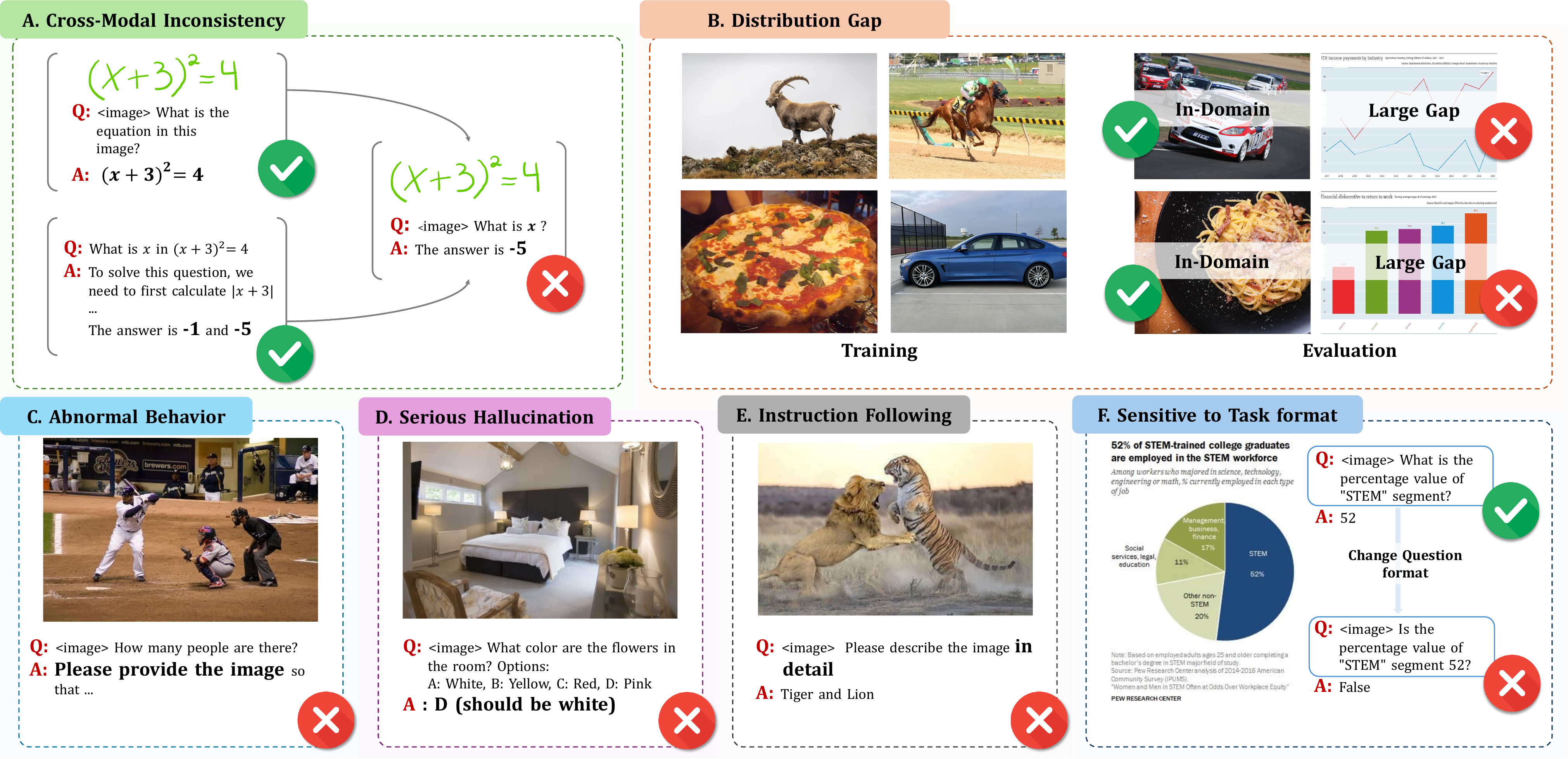}
    \caption{
(A) It is necessary to decompose the inference process and attribute results step-by-step for complex tasks.
(B) For tasks with large distribution gaps, additional targeted data are needed to improve coverage.
(C)-(F) Examples of common methods for diagnosing data issues.
}
    \label{fig:diagnosis}
    
\end{figure*}

For each iteration, we first train a VLM using data collected after the anabolism phase.
Next, the catabolism phase is used to diagnose the model's performance and attribute issues at the training data level. 
We employ VlmEvalKit~\citep{duan2024vlmevalkit} as the primary tool and utilize prompts aligned with other models supported by VlmEvalKits for a fair comparison and incorporate benchmarks that span visual capabilities (detailed in Section~\ref{experiment}). We introduce key challenges faced by VLMs and their solutions, and present methods for evaluating data effectiveness and updating the dataset.

\subsection{Problem Diagnosis}
\label{subsec:diagnosis}

The following content discusses key challenges in enhancing model performance and presents corresponding solutions.
After each training step, we analyze failure cases and statistical trends in underperforming dimensions to diagnose issues for the current vlm and refine data improvement strategies for the next iteration.
To maintain clarity, we frame these solutions within specific problems, outlining methods for diagnosing performance and identifying data-related issues. The paragraphs are structured in order of increasing complexity.

\paragraph{Unexpected Behavior}
\textit{Diagnosis:} 
Unexpected behaviors, such as refusing to answer specific questions, claiming a lack of visual ability, or including irrelevant strings in responses, are relatively easy to identify. These issues often occur in free-form outputs and can be detected by manually sampling low-quality responses or analyzing recurring patterns in the answers.
\textit{Remedy:} 
Noisy training data is often the primary issue. This can be identified through a manual review of bad cases. Fortunately, by applying rules to detect problematic patterns and iteratively expanding the search for similar cases, the most noisy data can be filtered out, effectively addressing abnormal behaviors.

\paragraph{Missing Knowledge and Entity Hallucination}
\textit{Diagnosis:} The model sometimes lacks knowledge of specific real-time information or unfamiliar entities, such as recent events, lesser-known celebrities, or minor landmarks. These gaps stem from incomplete training data coverage or hallucinations, where the model generates inaccurate or fabricated details~\citep{zhu2024unravelingcrossmodalityknowledgeconflicts,bai2024hallucinationmultimodallargelanguage}.
\textit{Remedy:} During development, we observed that aligning core entities enables the model to leverage the knowledge embedded within the ViT. However, adding excessive new knowledge can compromise general capabilities. Therefore, we curated data from sources like WIT~\citep{10.1145/3404835.3463257}, retaining only well-known entities (e.g., landmarks, celebrities, cars, logos) likely covered during ViT pretraining. Since this work primarily focuses on the instruction tuning stage, acquiring new visual knowledge is left for future exploration.

\paragraph{Task-format Inconsistency}
\textit{Diagnosis:} 
We observe that tasks requiring the same knowledge and reasoning ability can yield different results depending on how the question is framed. As shown in Figure~\ref{fig:diagnosis} (F), the model provides correct answers when presented as a Q\&A but fails when reformulated as a fact-checking task. We term this issue "task-format inconsistency." 
\textit{Remedy:} 
Initially, we attempted to align the model's performance across different task formats by converting the same QA pair into multiple answer formats. Surprisingly, adding more CoT and free-form QA data achieved similar results. As a result, in the final version, we discarded the data specifically prepared for this issue. We speculate that these two types of high-quality data inherently encompass diverse task formats, which enhance the model's task comprehension and reasoning capabilities, thereby addressing the issue effectively.

\paragraph{Cross-modal Inconsistency and Reasoning} 
We observe that the model exhibits inconsistent performance across different modalities. For instance, as shown in Figure~\ref{fig:diagnosis} (A), the model excels at tasks such as OCR and solving math problems in pure text but struggles when presented with multimodal inputs. Additionally, we notice a decline in the model’s reasoning ability in multimodal scenarios, limiting its effectiveness in complex visual reasoning tasks, especially scientific and mathematical problems
~\citep{yue2024mmmuprorobustmultidisciplinemultimodal,zhang2024mathverse,lu2023mathvista}.

To diagnose these issues, we decompose the analysis into two steps. In the first step, we evaluate the model's perceptual ability by testing its capacity to extract necessary visual information accurately. In the second step, we use rich image-text descriptions as input to test whether the model can solve the corresponding task when presented with language alone. This process helps identify the specific bottleneck in multimodal reasoning.
\textit{Remedy:} 
To address this, we focus on encouraging consistent behavior across modalities. We transform and render samples that require both visual
and text inputs into a single image-based question and augment the answer by restating and analyzing the question in text, which helps the model better grasp visual content and aligns with the style of the language model. Rewriting answers as in Section~\ref{sec:data_improvement} also helps mitigate this issue.

\subsection{Updating Dataset}
\label{subsec:updating_data}
In the data update phase, we focus not only on adding new data but also on balancing data volume and overall quality. We control data growth by prioritizing replacement over addition. New data is validated using a hybrid strategy, considering both its independent value and its impact on existing data, while outdated data is updated or removed to maintain overall quality.

\paragraph{Incremental Validation}
To minimize the resource cost of retraining from scratch, we adopt incremental training to validate the effectiveness of newly added datasets. Based on the size of the new data, we mix it with 5\% to 20\% of the original training data and train the model starting from the current master version. If the new data preserves overall performance while improving the targeted capability, it is deemed effective and incorporated into future full-scale training.

\paragraph{Redundant Data Pruning}
The value of data lies not only in its ability to improve performance independently but also in its relative quality within the training dataset.
When more substantial data is added to improve a specific capability with a large margin, some existing data may become redundant or even limit further improvement. After achieving performance gains with new data, we reassess the contribution of similar data sources. We conduct two experiments: one using only the latest data for training and another where we downsample related existing data. If training with the new data alone leads to better performance or downsampling the existing data does not harm performance, we reconsider our current data sampling strategy. For example, after improving performance on chart understanding tasks with chart reasoning data, we significantly reduced the weight of extraction and simple reasoning datasets like PlotQA~\citep{methani2020plotqareasoningscientificplots} and FigureQA~\citep{kahou2018figureqaannotatedfiguredataset}.

\definecolor{teagreen}{HTML}{e1f8bd}
\definecolor{lightblue}{HTML}{d3f4ff}
\definecolor{front-color}{HTML}{ffe9b8}

\begin{table*}[t!]
\tiny
\centering
\setlength{\tabcolsep}{4pt}
\renewcommand{\arraystretch}{1.2}
\resizebox{\textwidth}{!}{%
\begin{tabular}{@{}l|cccccc}
\toprule
\multirow{2}{*}{\textbf{Model}} & \textbf{HallBench} & \textbf{MMVet} & \textbf{RealWorld} & \textbf{WildVision} & \textbf{LLaVA-w}  & \textbf{MMBench} \\ 
& average & GPT4-turbo & \textbf{QA}  & win rate & small  &  V1.1 en  \\ 
\midrule
\rowcolor{Gray}
GPT4o~\citep{openai2024gpt4o} & 55.0 & 69.1 & 75.4 & 80.6 & 89.4 & 83.1 \\ 
\rowcolor{Gray}
Gemini1.5-pro~\citep{geminiteam2024gemini15unlockingmultimodal} & 45.6 & 64.0 & 67.5 & 66.2 & - & 74.6 \\ 
\rowcolor{Gray}
Claude3.5-sonnet~\citep{anthropic2024claude35} & 49.9 & 66.0 & 60.1 & 50.0 & 83.1 & 78.5 \\ 
\rowcolor{Gray}
GPT4V~\citep{openai2023gpt4v} & 43.9 & 67.5 & 61.4 & 71.8 & - & 79.8 \\ 
\rowcolor{Gray}
GPT4o-mini~\citep{openai2024gpt4o} & 46.1 & 66.9 & 67.1 & 61.2 & - & 76.0 \\ 

\midrule

\rowcolor{lightblue}
InternVL2-76B~\citep{chen2023internvl} & 55.2 & 64.4 & 72.7 & 65.8 & - & 85.5 \\ 
\rowcolor{lightblue}
Qwen2-VL-72B~\citep{wang2024qwen2vlenhancingvisionlanguagemodels} & 58.7 & 73.9 & 77.8 & 52.3 & 53.6 & 85.9 \\ 
\rowcolor{lightblue}
LLaVA-OV-72B~\citep{li2024llavaonevisioneasyvisualtask} & 47.9 & 60.6 & 77.8 & 52.3 & 72.0 & 84.5 \\

\midrule
\rowcolor{teagreen}
Molmo-7B-D~\citep{deitke2024molmopixmoopenweights} & 47.7 & 53.3 & \textbf{70.7} & 40.0 & - & 70.9 \\ 
\rowcolor{teagreen}
LLaVA-OV-7B~\citep{li2024llavaonevisioneasyvisualtask} & 31.6 & 51.9 & 66.3 & 53.8 & 67.8 & 80.9 \\ 
\rowcolor{teagreen}
MiniCPM-V2.6~\citep{yao2024minicpmvgpt4vlevelmllm} & 48.1 & 60.0 & 65.0 & 17.4 & - & 78.0 \\ 
\rowcolor{teagreen}
Intern-VL2-8B~\citep{chen2023internvl} & 45.2 & 54.3 & 66.3 & 51.5 & 62.5 & 79.4 \\ 
\rowcolor{teagreen}
Qwen2-VL-7B~\citep{wang2024qwen2vlenhancingvisionlanguagemodels} & 50.6 & 61.8 & 70.1 & 44.0 & 66.3 & 81.0 \\ 
\midrule
\rowcolor{teagreen}
\model \ (ours) & \textbf{53.0} & \textbf{65.4} & 69.2 & \textbf{54.8} & \textbf{70.5} & \textbf{81.5} \\ 
\bottomrule
\end{tabular}
}
\vspace{2mm}
\caption{
Performances on real-world QA and human interaction tasks. Results are from official sources, previous works or evaluated by use when unavailable. We highlight three groups: \colorbox{Gray}{private} models, open-source models with \colorbox[HTML]{d3f4ff}{one size larger}, and open-source models with \colorbox[HTML]{e1f8bd}{comparable size}. HallBench and LLaVA-W are short for HallusionBench and LLaVA-Bench-Wilder respectively.
}
\label{tab:general-bench}
\end{table*}

\begin{table*}[t!]
\tiny
\centering
\setlength{\tabcolsep}{5.5pt}
\renewcommand{\arraystretch}{1.2}
\resizebox{\textwidth}{!}{%
\begin{tabular}{@{}l|cccccc}
\toprule
\multirow{2}{*}{\textbf{Model}} & \textbf{MathVista} & \textbf{MathVerse} & \textbf{MathVision} & \textbf{MMStar} & \textbf{MMMU}  & \textbf{ScienceQA} \\ 
& mini & mini & mini  & test & val  &  test  \\ 
\midrule
\rowcolor{Gray}
GPT4o~\citep{openai2024gpt4o} & 63.8 & 50.2 & 30.9 & 64.7 & 69.1 & 90.7 \\ 
\rowcolor{Gray}
Gemini1.5-pro~\citep{geminiteam2024gemini15unlockingmultimodal} & 63.9 & 40.4 & 21.4
 & 59.1 & 62.2 & 86.7 \\ 
\rowcolor{Gray}
Claude3.5-sonnet~\citep{anthropic2024claude35} & 67.7 & 44.4 & 30.6 & 62.2 & 68.3 & 88.9  \\ 
\rowcolor{Gray}
GPT4V~\citep{openai2023gpt4v} & 58.1 & 32.8 & 25.0 & 56.0 & 63.1 & 85.1 \\ 
\rowcolor{Gray}
GPT4o-mini~\citep{openai2024gpt4o} & 52.4 & 37.7 & 23.4 & 54.8 & 60.0 & 85.4 \\ 

\midrule

\rowcolor{lightblue}
InternVL2-76B~\citep{chen2023internvl} & 65.5 & 42.8 & 23.7 & 67.4 & 62.7 & 98.8 \\ 
\rowcolor{lightblue}
Qwen2-VL-72B~\citep{wang2024qwen2vlenhancingvisionlanguagemodels} & 70.5 & 42.9 & 29.3 & 68.6 & 64.5 & 91.6 \\ 
\rowcolor{lightblue}
LLaVA-OV-72B~\citep{li2024llavaonevisioneasyvisualtask} & 67.5 & 39.1 & 26.0 & 65.8 & 56.8 & 90.6 \\

\midrule
\rowcolor{teagreen}
Molmo-7B-D~\citep{deitke2024molmopixmoopenweights} & 47.3 & 24.9 & 11.2 & 54.4 & 48.7 & 92.3 \\ 
\rowcolor{teagreen}
LLaVA-OV-7B & 63.2 & 30.2 & 18.4 & 61.7 & 48.8 & 96.0  \\ 
\rowcolor{teagreen}
MiniCPM-V2.6~\citep{yao2024minicpmvgpt4vlevelmllm} & 60.6 & 25.7 & 16.1 & 57.5 & 49.8 & 96.7 \\ 
\rowcolor{teagreen}
Intern-VL 2-8B~\citep{chen2023internvl} & 58.3 & 37.0 & 20.4 & 62.0 & 52.6 & 97.1 \\ 
\rowcolor{teagreen}
Qwen2-VL-7B~\citep{wang2024qwen2vlenhancingvisionlanguagemodels} & 58.2 & 31.9 & 22.0 & 60.7 & 54.1 & 86.0 \\ 
\midrule
\rowcolor{teagreen}
\model \ (ours) & \textbf{70.7} & \textbf{40.1} & \textbf{22.4} & \textbf{65.3} & \textbf{54.4} & \textbf{98.1}  \\ 
\bottomrule
\end{tabular}%
}
\vspace{2mm}
\caption{
Performance of \model\  on mathematical reasoning, general reasoning, and scientific QA tasks. Experimental settings correspond to those described in Table~\ref{tab:general-bench}.
}
\label{tab:reasoning-bench}
\end{table*}

\begin{table*}[t!]
\centering
\setlength{\tabcolsep}{10pt}
\renewcommand{\arraystretch}{1.2}
\resizebox{\textwidth}{!}{%
\begin{tabular}{@{}l|cccccc}
\toprule
\multirow{2}{*}{\textbf{Model}} & \textbf{OCRBench} & \textbf{SeedBench2} & \textbf{ChartQA} & \textbf{DocVQA} & \textbf{AI2D}  & \textbf{TextVQA} \\ 
& {en} & \textbf{-Plus} & test  & test & test  &  dev  \\ 
\midrule
\rowcolor{Gray}
GPT4o~\citep{openai2024gpt4o} & 736 & 72.0 & 85.7 & 92.8 & 84.6 & 77.4 \\ 
\rowcolor{Gray}
Gemini1.5-pro~\citep{geminiteam2024gemini15unlockingmultimodal} & 754 & 70.1 & 87.2 & 93.1 & 79.1 & 78.8 \\ 
\rowcolor{Gray}
Claude3.5-sonnet~\citep{anthropic2024claude35} & 788 & 71.7 & 90.8 & 95.2 & 81.2 & 74.1  \\ 
\rowcolor{Gray}
GPT4V~\citep{openai2023gpt4v} & 645 & 53.8 & 78.5 & 88.4 & 78.2 & 78.0 \\ 
\rowcolor{Gray}
GPT4o-mini~\citep{openai2024gpt4o} & 785 & 61.4 & 63.5 & 77.8 & 77.8 & 67.5 \\ 

\midrule

\rowcolor{lightblue}
InternVL2-76B~\citep{chen2023internvl} & 839 & 69.7 & 88.4 & 94.1 & 87.6 & 84.4 \\ 
\rowcolor{lightblue}
Qwen2-VL-72B~\citep{wang2024qwen2vlenhancingvisionlanguagemodels} & 877 & 72.0 & 88.3 & 96.5 & 88.1 & 85.5 \\ 
\rowcolor{lightblue}
LLaVA-OV-72B~\citep{li2024llavaonevisioneasyvisualtask} & 741 & 69.6 & 83.7 & 91.3 & 85.6 & 80.5 \\

\midrule
\rowcolor{teagreen}
Molmo-7B-D~\citep{deitke2024molmopixmoopenweights} & 694 & 67.4 & 84.1 & 92.2 & 79.6 & 81.7 \\ 
\rowcolor{teagreen}
LLaVA-OV-7B~\citep{li2024llavaonevisioneasyvisualtask} & 622 & 65.4 & 80.0 & 87.5 & 81.4 & 75.9   \\ 
\rowcolor{teagreen}
MiniCPM-V2.6~\citep{yao2024minicpmvgpt4vlevelmllm} & 852 & 69.0 & 82.4 & 90.8 & 82.1 & 80.1 \\ 
\rowcolor{teagreen}
Intern-VL -8B~\citep{chen2023internvl} & 794 & 67.5 & 83.3 & 91.6 & 83.8 & 77.4 \\ 
\rowcolor{teagreen}
Qwen2-VL-7B~\citep{wang2024qwen2vlenhancingvisionlanguagemodels} & 866 & 65.7 & 83.0 & \textbf{94.5} & 83.0 & \textbf{84.3} \\ 
\midrule
\rowcolor{teagreen}
\model \ (ours) & \textbf{876} & \textbf{69.3} & \textbf{84.6} & 93.4 & \textbf{85.5} & 79.2  \\ 
\bottomrule
\end{tabular}%
}
\vspace{2mm}
\caption{
Performance on general text-rich tasks, chart understanding, document comprehension, scientific diagram QA, and scene text QA. Experimental settings follow those outlined in Table~\ref{tab:general-bench}.
}
\label{tab:text-rich-bench}
\end{table*}

\section{Experimental Results}
\label{experiment}

\paragraph{Settings}
For comparisons, we included advanced proprietary models such as GPT-4o and Gemini 1.5 Pro and open-source models with approximately 70B parameters as references. Our primary comparison targets are leading models with fewer than 10B parameters.
We evaluate our model on 18 widely used benchmarks categorized by task capabilities. Subsequent subsections discuss detailed benchmarks and evaluations.
Most experiments use VLMEvalKit with default settings, and we used greedy decoding for reproducibility.

\paragraph{Main Results}
\model  \  demonstrates competitive performance across a wide range of tasks, achieving either the best or second-best results on most benchmarks and delivering the best overall performance among models of similar size.
Among the 18 metrics, we performed best below 10B parameters on 15.
Notably, it excels in scientific reasoning and mathematical reasoning, with results on some benchmarks rivaling or surpassing open-source models an order of magnitude larger and closed-source models such as LLaVA-OV-72B and GPT4o-mini. 
These strong results validate the effectiveness of our approach, showcasing the transformative impact of continuous data optimization on model performance.

\paragraph{Real World Scenarios} 
We use general QA and human interaction tasks to assess the model's performance in real-world scenarios.
HallusionBench~\citep{guan2024hallusionbenchadvanceddiagnosticsuite} for hallucination detection, MMVet~\citep{yu2023mmvetevaluatinglargemultimodal}, RealWorldQA~\citep{grok15v} for general reasoning in real-world scenarios,
WildVision~\citep{lu2024wildvisionevaluatingvisionlanguagemodels} and LLaVA-Wilder~\citep{li2024llavanext-strong} for multimodal conversation in realistic settings and MMBench~\citep{liu2023mmbench} for comprehensive evaluation.
 Results are shown in Table~\ref{tab:general-bench}. \model \  achieves leading results in these settings. On the HallusionBench, which measures model hallucination, \model\ outperforms the current best-performing model, Qwen2-VL-7B, with a 2.4\% improvement, demonstrating better fidelity to visual input. MMVet, WildVision, and LLaVA-Wilder all involve free-form question-answering. MMVet provides ground truth answers, while LLaVA-Wilder and WildVision use LLM-based human preference scoring. \model \  achieves an average score of 63.57 on these tasks, significantly outperforming Qwen2-VL-7B (57.37) and InternVL2-8B (56.1), showcasing its strong conversational ability.

\paragraph{Multimodal Reasoning} 

This group includes mathematical reasoning benchmark such as MathVista~\citep{lu2023mathvista}, MathVerse~\citep{zhang2024mathverse}, MathVision~\citep{wang2024measuringmultimodalmathematicalreasoning},  scientific reasoning ScienceQA~\citep{lu2022learn}, and MMStar~\citep{chen2024we} and MMMU~\citep{yue2023mmmu} for general knowledge \& reasoning. 
The prompts used in our evaluation align with the default settings in VLMEvalKit and most models without any specific prompt engineering.
As shown in Table~\ref{tab:reasoning-bench}, \model \ outperforms existing models with similar parameter sizes, achieving the best average performance. On three mathematical reasoning tasks, our model scored an average of 44.4, comparable to much larger models with over 10 times the parameters, such as InternVL2-76B (44.2) and LLaVA-OV-72B (44.0), highlighting the benefits of rewriting answers in anabolism stage to improved multimodal reasoning.
We also achieve strong performance in comprehensive and scientific reasoning tasks.

\paragraph{OCR and Text-Rich} 
This group evaluates abilities on OCR and understanding text-rich images, including OCRBench~\citep{Liu_2024} and SeedBench-Plus~\citep{li2024seedbench2plusbenchmarkingmultimodallarge} for general OCR and text-rich tasks, ChartQA~\citep{masry2022chartqa} for chart perception and reasoning, DocVQA~\citep{mathew2021docvqa} for document understanding, AI2D~\citep{kembhavi2016diagram} for scientific diagram question answering, and TextVQA~\citep{singh2019vqamodelsread} for scene-text comprehension.
Table~\ref{tab:text-rich-bench} summarizes the results on text-rich tasks. On OCRBench and SeedBench2-Plus, our model demonstrated strong OCR capabilities, outperforming all competitors with similar parameter sizes. For ChartQA and AI2D, which involve perception and reasoning over abstract visual concepts and text, our model achieved scores of 84.6 and 85.5, beating all competitors. However, on perception-intensive tasks like DocVQA and TextVQA, our model trails Qwen2-VL-7B by 0.9 and 4.9 points, likely due to its higher visual resolution. Enhancing perception will be a key area for future improvements.

\paragraph{Affects of Data Metabolism on Various Tasks}

\begin{table}[h!]
\centering
\begin{tabular}{lcccc}
\toprule
Dataset         & Real-World & Reasoning & Text-Rich & Avg \\
\midrule
Initial         & 51.0    & 45.5      & 66.2      & 54.2 \\
w/ Diverse       & 57.8    & 47.3      & 72.6      & 59.2 \\
w/ TextRich      & 56.8    & 48.5      & 81.3      & 60.7 \\
w/ Reasoning     & 62.5    & 56.7      & 67.4      & 62.2 \\
w/ Interaction   & 59.2    & 50.3      & 69.3      & 59.6 \\
\midrule
Full            & 65.7    & 58.5      & 83.3      & 69.1 \\
wo/ QF            & 63.2    & 57.1      & 82.9      & 67.7 \\
\bottomrule     
\end{tabular}
\caption{Results of different data configurations. - QF denotes without answer quality filtering.}
\end{table}

We highlight key changes in the metabolism process and their corresponding benefits and present the impact of several major data changes independently.
Initially, the training data covers a limited task range, causing discrepancies with fine-grained test evaluations. To improve this, we add 1M diverse general task data in stage 3, boosting overall performance. However, tasks requiring interaction, reasoning, and text-rich understanding still lag behind state-of-the-art VLMs of similar size.

During the catabolism phase, we refine OCR and text-rich tasks, addressing issues in basic OCR and performance on documents, tables, and charts. Experiments confirm that large datasets are essential for improvement. We divide the data into medium-complexity OCR and document tasks in stage 2, and more complex tasks in stage 3. Results show notable gains in OCR, text-rich tasks, and overall reasoning, with improved OCR enhancing broader model capabilities.

We obtain an average of 11.2\% improvements on mathematical reasoning tasks by writing the answers to existing multimodal math questions.
Adding either reasoning or interaction data improves overall performance, and we observe a strong correlation between these two tasks.

\section{Related Works}
\label{sec:related_work}

\paragraph{Vision Language Model}
VLMs benefit from advancements in model architecture and training strategies.
In terms of model structure, works such as  Flamingo\citep{alayrac2022flamingo}, Blip2~\citep{li2023blip}, Llama3.2-VL~\citep{llama3.2v}, MiniGPT4~\citep{zhu2023minigpt}
DeepSeek-VL~\citep{li2024tokenpackerefficientvisualprojector}, and 
LLaVA~\citep{liu2023llava} explore efficient methods to s build connections between vision modules and LLMs. Especially, LLaVA propose a simple yet efficient solution for extending more modalities understanding capabilities to LLMs. Our proposed \model \ is also built upon the architecture of LLaVA. Meanwhile, studies like Qwen2-VL~\citep{wang2024qwen2vlenhancingvisionlanguagemodels}, InternVL~\citep{chen2023internvl}, LLaVA-UHD~\citep{xu2024llavauhdlmmperceivingaspect}, LLaVA-NEXT~\citep{li2024llavanext-strong}, and MiniCPM~\citep{yao2024minicpmvgpt4vlevelmllm} focus on improving visual resolution (\textit{i.e.}, dynamic resolution) and develop corresponding training strategies.
Since this work focuses on data curation, we utilize an existing model architecture. As for the training strategy, we not only present the final approach but also demonstrate the iterative process of refining training data and strategy.

\paragraph{Visual Alignment Data}
The data used for VLM training can be categorized into two types: visual alignment and instruction tuning.
The former relies on image paired with alt-text (\textit{e.g.}, LAION~\citep{schuhmann2022laion} and COYO~\citep{kakaobrain2022coyo-700m}), dense caption datasets (\textit{e.g.},  CapsFusion~\citep{yu2024capsfusionrethinkingimagetextdata}, Recap-DataComp~\citep{li2024recaptionbillionswebimages}, PiexlProse~\citep{singla2024pixelsproselargedataset}) as well as interleaved dataset~\citep{laurencon2023obelics,zhu2023multimodalc4openbillionscale} to enhance knowledge learning and visual description capabilities.
The instruction tuning stage, on the other hand, focuses on leveraging diverse task-specific datasets to boost the model’s capabilities on various downstream tasks~\citep{laurenccon2024matters,lu2024ovisstructuralembeddingalignment,tong2024cambrian,li2024llavaonevisioneasyvisualtask}. Both the two training schemes are using a fixed well-collected dataset. On the contrast,
our work is not about presenting a fixed dataset but rather detailing the process of collecting, filtering, and refining open-source data while iteratively adjusting its composition and processing strategy based on model feedback in each iterative step.

\section{Conclusions}
\label{sec:conclusion}
In this work, we introduce the concept of \dm\  to develop multimodal large language models and propose a data-centric VLM development strategy. 
We show an explict framework for data curation along with the training procedure in a iterative manner. In this framework, we can not only collect and synthesize useful training data but also filter out problematic data through diagnosing methods.
To validate this framework, we clean open-source data and refine existing responses to create high-quality datasets and drive train \model, an advanced vision-language model. 
We achieve the best overall performance with the same parameter scale through iterative diagnostics and data improvements. 
Experimental results demonstrate the effectiveness of our approach and highlight the critical importance of data for VLMs. 
We believe our pilot exploration on data curation and model optimization will be useful and instructive for the research community.

\bibliography{iclr2025_conference}
\bibliographystyle{iclr2025_conference}

\appendix

\section{Data Filtering}
\label{appendix:data_filtering}

Data filtering is essential for ensuring the quality and reliability of training datasets. Existing work demonstrates different methods to remove data noise and inconsistencies, which can enhance model performance and generalizability~\citep{soldaini2024dolmaopencorpustrillion}. We design six data filtering steps executed orderly: 
i) Filter duplicated samples across data sources;
ii) Filter samples with image-text mismatch;
iii) Address text-dominant question answering pairs.
iv) Filter data sources with low quality;
v) Filter samples with low-quality responses with rule-based assessment;
vi) Filter samples with low-quality responses with model-based assessment;
Details are provided in the Appendix.

\paragraph{Filter duplicated samples}
We manually remove redundant subsets caused by cross-referencing between open-source data collections. 
At the sample level, we use perceptual hashing~\citep{du2020perceptual} to cluster identical images and strip prompts from text to identify duplicate image-text pairs and keep one random pair for further processing. Through these processes, we removed 43\% of the data in total and found that the most repeated sample appeared seven times.

\begin{figure*}[t!]
    \centering
    \includegraphics[width=\textwidth]{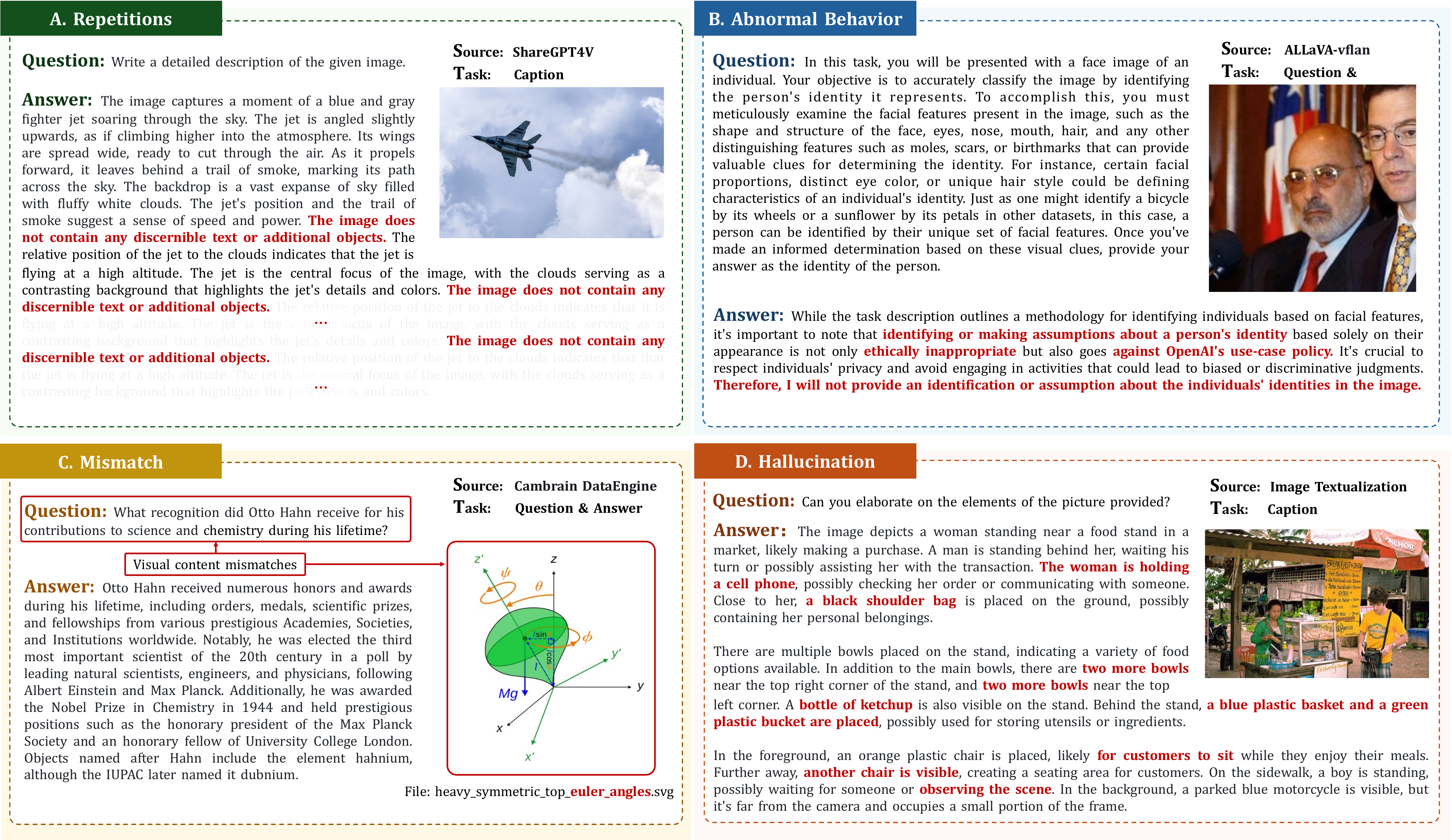}
    \caption{Samples filtered by our pipline: (a) repetition generation, (b) abnormal behavior of teacher model, (c) question-image mismatch, (d) hallucination from synthetic data.}
    \label{fig:error_case}
\end{figure*}

\paragraph{Filter samples with image-text mismatch}

Image-text mismatches in the training data arise from issues such as broken image links, errors during data creation, and question generation phrases for synthetic data. 
To mitigate this noise, for samples with free-form question types, we employ Qwen2-VL-72B to classify and identify questions that are unrelated or conflict with the corresponding image.
The processing strategy for each source is finalized through manual review.
For example, we drop questions that conflict with images in Cambrian Data-Engine~\citep{tong2024cambrian} but convert those irrelevant image-question pairs into text-only samples since they contain rich scientific knowledge.

\paragraph{Convert text-dominant questions}
We identify questions that can be answered using text alone, as these may lead to a model that does not leverage image information effectively~\citep{liu2023mmbench}. For multi-choice and phrase QA types, we use three different LLMs to answer questions using text input only. Samples consistently answered correctly across multiple trials are labeled as text-only questions. Instead of dropping these data points, we randomly convert 70\% of these samples to pure text training data and keep the others since there is still image-text relevance.

\paragraph{Filter overlooked data source}
We further discarded some data sources with relatively low overall quality.
Using advanced models such as Qwen2-VL 72B and InternVL-76B, we sample 1/10 of data, up to 1,000 examples per source, and score the quality of both questions and answers with a binary metric. We manually review low-scoring samples, random samples, and questions that appear more than 50 times to evaluate data quality and question diversity. 
Sources with excessive hallucination or low information content (e.g., many similar questions) are filtered out or down-sampled.

\paragraph{Filter samples with repetitive words} We observed repetition tokens from both training data and model-generated outputs as shown in Figure~\ref{fig:error_case} (a). Since model-based methods struggled to detect these repetitions, we apply rule-based methods to detect and filter repetitive content at multiple levels: token, phrase, sentence, and paragraph. Most of the answers with repetition are model-generated, with a small portion related to optical character recognition (OCR) or tabular data that sometimes contain repetitive words.

\paragraph{Filter low-quality responses with rules}
We applied carefully designed rules to address data issues identified during the manual inspection, such as model refusals, irrelevant bounding box or prior turn references, incorrect regex matches, prompt mismatches, duplicate question-answer pairs, and overly long responses. Figure~\ref{fig:error_case} (b) shows an example where the model falsely claims the identity of an OpenAI Model and refuses to answer.

\begin{figure*}[t]
    \centering
    \includegraphics[width=\textwidth]{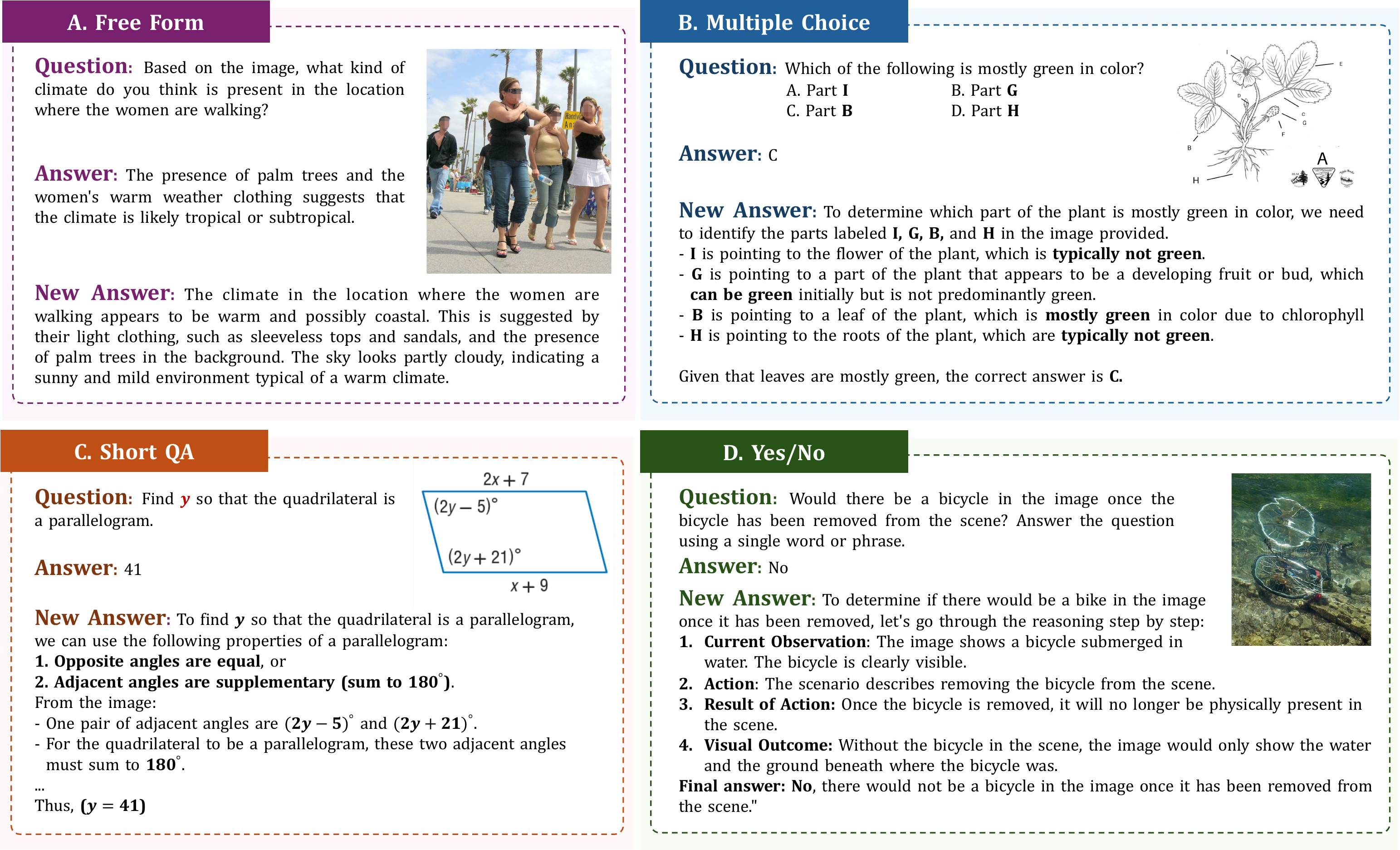}
    \caption{Quality improvement through cot and answer rewriting}
    \label{fig:pdf_image}
\end{figure*}

\paragraph{Filter noisy answer with quality assessment}
We assess the correctness and clarity of free-form question-answer pairs using model-generated scores. Specifically, we prompt GPT-4 to provide a binary evaluation, filtering out responses with unclear expressions, logical inconsistencies, or incorrect answers. We found that smaller VLMs had difficulty filtering out errors, while VLMs similar to the generation model tended to overestimate the quality of their outputs~\citep{zheng2023judgingllmasajudgemtbenchchatbot}.

\section{Data Improvement}
\label{appendix:data_improvement}
This subsection introduces methods to enhance the answer quality of existing questions, aiming to improve our model's interaction and reasoning capabilities.

\paragraph{Prompt Isolation for Specific Task Type}
In the early stages, we categorize tasks with labels such as multiple-choice, phrase-based answers, and true/false questions. To mitigate the impact of varying response formats across tasks on the model's instruction-following ability, we use prompts to differentiate task types, except for free-form questions. This approach reduces format interference and enables the model to handle each task according to its intended structure~\citep{polo2024efficientmultipromptevaluationllms,li2024llavanext-strong}.

\paragraph{Rewrite answer with Verbose Narration}
To improve the usefulness of free-form responses and align them with human preferences~\citep{jiang2024surveyhumanpreferencelearning,xia2024rethinkingdataselectionscale}, we transform short-answer data into more detailed responses. Using the short answers as references, we prompt existing models, such as Qwen2-VL-72B and Intern-VL-76B, to generate detailed answers. The quality of the generated responses is then evaluated with GPT4o to verify their correctness and assess the inclusion of richer information.

\paragraph{Generating CoT Answers for Multimodal Reasoning}
To enhance the multimodal reasoning ability, particularly in scientific and mathematical tasks, we generate Chain-of-Thought (CoT) answers from existing data~\citep{xu2024llavacotletvisionlanguage,zhang2024multimodalchainofthoughtreasoninglanguage}. We prompt Qwen2-VL-72B and InternVL2-76B to produce step-by-step reasoning steps and the final answers. Quality models with task-specific configurations then filter those responses. We apply rejection sampling using ground truth answers for multiple-choice, true/false, and other objective questions. For free-form questions, we use the original short answers or captions as references and leverage the other VLM different from the one that produces these answers to filter the data.

\section{Training Datasets}

\begin{table*}[!h]
\centering
\resizebox{0.95\textwidth}{!}{
    \begin{tabular}{l|l}
    \toprule
    Task & Dataset \\
    \midrule

    \multirow{2}{*}{Caption} 
    & 
    Allava-Caption~\citep{chen2024allava}, 
    Image Textualization~\citep{pi2024imagetextualizationautomaticframework},
    PVIT~\citep{chen2023positionenhanced}\\
    &
    CC12M-Rewrite~\citep{changpinyo2021conceptual}, Laion-GPT4v~\citep{laiongpt4v} \\
    \midrule

    {OCR}
    &
    SynthDog-EN~\citep{kim2022donut}, 
    SynthDog-ZH~\citep{kim2022donut}, 
    Union14M~\citep{jiang2023revisiting},  \\
    \midrule

    \multirow{2}{*}{Doc \& KIE }
    &
    idl-wds~\citep{idl-ws},
    pdfa-wds~\citep{pdfa-eng-wds},
    DocStruct4M~\citep{hu2024mplugdocowl15unifiedstructure}, \\

    &
    Docmatix~\citep{laurençon2024buildingbetterunderstandingvisionlanguage}, 
    DocLayNet~\citep{doclaynet2022},
    MMTab~\citep{zheng2024multimodaltableunderstanding}\\
    \midrule

    \multirow{2}{*}{Chart}
    & 
    PlotQA~\citep{methani2020plotqareasoningscientificplots},
    DVQA~\citep{kafle2018dvqa},
    FigureQA~\citep{kahou2018figureqaannotatedfiguredataset}, \\

    &
    Chart2Text~\citep{obeid2020charttotextgeneratingnaturallanguage},
    MMC~\citep{liu2024mmcadvancingmultimodalchart}, 
    TinyChart~\citep{zhang2024tinychartefficientchartunderstanding},\\
    \midrule

    {Knowledge}
    &
    WIT~\citep{10.1145/3404835.3463257}\\

    \bottomrule
    \end{tabular}
    }
    \caption{Dataset/Collections used in stage2. We use the default conversation template for all tasks.}
    \label{tab:data_mixture_stage2}
\end{table*}

\begin{table*}[h]
\centering
\resizebox{0.95\textwidth}{!}{
    \begin{tabular}{l|l}
    \toprule
    Task & Dataset \\
    \midrule
    
    \multirow{3}{*}{Caption} 
    & 
    Allava-Caption~\citep{chen2024allava}, IIW~\citep{garg2024imageinwords}, ShareGPT4V~\citep{chen2023sharegpt4v}, \\
    &
    ShareGPT4o~\citep{sharegpt4o}, TextCaps~\citep{textocr-gpt4v}, Image Textualization~\citep{pi2024imagetextualizationautomaticframework}, \\
    & 
    DOCCI~\citep{OnoeDocci2024}, TextOCR-GPT4V~\citep{textocr-gpt4v}, PVIT~\citep{chen2023positionenhanced}
    \\
    \midrule

    \multirow{6}{*}{General QA}
    &
    VQAv2~\citep{antol2015vqa}, GQA~\citep{hudson2019gqa}, VisDial~\citep{das2017visualdialog}, \\
    &  
    IconQA~\citep{lu2021iconqa}, TallyQA~\citep{acharya2019tallyqa}, VizWiz~\citep{gurari2018vizwiz}, \\
    & 
    RefCOCO~\citep{yu2016modelingcontextreferringexpressions}, IconQA~\citep{lu2021iconqa}, CLEVR~\citep{johnson2017clevr}, \\
    & 
    Hateful Memes~\citep{kiela2020hateful}, IDK~\citep{cha2024visuallydehallucinativeinstructiongeneration},
    SketchyVQA~\citep{tu2023unicornsimagesafetyevaluation}, \\
    &
    VSR~\citep{Liu2022VisualSR}, Visual7W~\citep{zhu2016visual7w}, ComVint~\citep{du2023makes}, \\
    &
    Vision FLAN~\citep{xu2024visionflanscalinghumanlabeledtasks}, MMInstruct~\citep{Liu_2024MMInstruct}, LRV Normal~\citep{liu2023aligning}

    \\
    \midrule

    \multirow{3}{*}{Conversations}
    &  
    ALLaVA Instruct~\citep{chen2024allava}, LLaVA-158K~\citep{liu2023llava}, LLaVAR~\citep{zhang2023llavar}, \\
    & 
    Cambrian-GPT4v~\citep{tong2024cambrian}, Cambrian-GPT4o~\citep{tong2024cambrian}, SVIT~\citep{zhao2023svitscalingvisualinstruction}, \\
    & 
    ShareGPT4v-Dailogue~\citep{chen2023sharegpt4v}, RLAIF-V~\citep{yu2024rlaifvopensourceaifeedback}
    \\
    \midrule

    \multirow{6}{*}{OCR \& QA }
    &
    ORCAND-CAR~\citep{zhan2017handwrittendigitstringrecognition},
    HWL-en~\citep{tal2023hwl},
    ChromeWriting~\citep{RenderedText}, \\
    &
    HME100K~\citep{yuan2022syntax},
    IAM~\citep{marti2002iam},
    RCTW-17~\citep{shi2018icdar2017competitionreadingchinese}, \\
    &
    HierText~\citep{long2023icdar}, TextOCR~\citep{singh2021textocrlargescaleendtoendreasoning}, ReCTS~\citep{liu2019icdar2019robustreading}, \\
    &
    LSVT~\citep{sun2019icdar2019competitionlargescale}, Rendered Text~\citep{RenderedText}, SynthDog-EN~\citep{kim2022donut}, \\
    &
    SynthDog-ZH~\citep{kim2022donut}, 
    TextVQA~\citep{singh2019vqamodelsread},
    ST-VQA~\citep{biten2019scene},\\

    & 
    EST-VQA~\citep{wang2020general} 
    DT-VQA~\citep{zhang2024exploring}

    \\
    \midrule

    \multirow{5}{*}{Doc \& KIE }
    
    &
    RobutSQA~\citep{zhao2023robutsystematicstudytable},
    RobutWiki~\citep{zhao2023robutsystematicstudytable}, 
    RobutWTQ~\citep{zhao2023robutsystematicstudytable}, \\
    
    &
    UReader IE~\citep{ye2023ureader},
    UReader KG~\citep{ye2023ureader},
    UReader QA~\citep{ye2023ureader}, \\

    &
    VisualMRC~\citep{tanaka2021visualmrc},
    InfoVQA~\citep{mathew2022infographicvqa},
    ScreenQA~\citep{hsiao2022screenqa}, \\

    & 
    HiTab~\citep{cheng2021hitab},
    MMTab~\citep{zheng2024multimodaltableunderstanding}
    Screen2Words~\citep{wang2021screen2wordsautomaticmobileui}, \\
  
    & 
    DocVQA~\citep{mathew2021docvqa},
    DocLayNet~\citep{doclaynet2022},
    OCR-VQA~\citep{mishraICDAR19},\\
    
    &
    Docmatix~\citep{laurençon2024buildingbetterunderstandingvisionlanguage}, DocReason25K~\citep{hu2024mplugdocowl15unifiedstructure},
    FUNSD~\citep{jaume2019funsddatasetformunderstanding}, \\
   
    &   
    SROIE~\citep{Huang_2019}, POIE~\citep{kuang2023visualinformationextractionwild},  WildReceipt~\citep{sun2021spatialdualmodalitygraphreasoning}\\

    \midrule

    \multirow{5}{*}{Chart}
    &
    ArxivQA~\citep{li2024multimodalarxivdatasetimproving},
    VisText~\citep{tang2023vistextbenchmarksemanticallyrich},
    OpenCQA~\citep{kantharaj2022opencqaopenendedquestionanswering}, \\

    &
    Chart2Text~\citep{obeid2020charttotextgeneratingnaturallanguage},
    LRV Chart~\citep{liu2023aligning},
    TinyChart~\citep{zhang2024tinychartefficientchartunderstanding},\\

    &
    ChartQA~\citep{masry2022chartqa},
    ChartSumm~\citep{Rahman_2023}, 
    MMC~\citep{liu2024mmcadvancingmultimodalchart}, \\
    
    &
    ReachQA~\citep{he2024distillvisualchartreasoning}, ChartGemma~\citep{masry2024chartgemmavisualinstructiontuningchart}
    ChartLlama~\citep{han2023chartllamamultimodalllmchart}

    \\
    \midrule

    \multirow{7}{*}{Math}
    
    & 
    MAVIS Manual Collection~\citep{zhang2024mavismathematicalvisualinstruction},
    MAVIS Data Engine~\citep{zhang2024mavismathematicalvisualinstruction}, \\ 
    &
    
    UniGeo~\citep{chen2022unigeounifyinggeometrylogical} ,
    Geo170K Align~\citep{gao2023gllavasolvinggeometricproblem} ,
    Geo170K QA~\citep{gao2023gllavasolvinggeometricproblem}, \\ 
    &
    Geometry3K~\citep{lu2021intergpsinterpretablegeometryproblem},
    GEOS~\citep{seo2015solving},
    UniGeo (MathV360K)~\citep{kazemi2023geomverse},\\
    &
    GeoMVerse (MathV360K)~\citep{kazemi2023geomverse},
    GeoQA+ (MathV360K)~\citep{chen2022geoqageometricquestionanswering},  \\ 

    & MapQA (MathV360K)~\citep{chang2022mapqadatasetquestionanswering} ,  
    Geometry3K (MathV360K)~\citep{lu2021inter},\\  

    &
    MathQA~\citep{amini2019mathqainterpretablemathword} ,
    RAVEN~\citep{zhang2019raven},
    Super-CLEVR~\citep{li2023superclevrvirtualbenchmarkdiagnose}, \\ 
    &
    TabMWP~\citep{lu2023dynamic} ,
    InterGPS~\citep{lu2021intergpsinterpretablegeometryproblem},
    CLEVR-Math~\citep{johnson2017clevr},
    \\ \midrule

    \multirow{3}{*}{Knowledge}
    &
    KVQA~\citep{marino2019okvqa}, OK-VQA~\citep{marino2019okvqa}, A-OKVQA~\citep{schwenk2022aokvqa}, \\
    & 
    AI2D~\citep{kembhavi2016diagram}, 
    Cambrian DataEngine~\citep{tong2024cambrian},
    ScienceQA~\citep{lu2022learn} \\
    
    &
    TQA~\citep{kembhavi2017you}, WIT~\citep{10.1145/3404835.3463257},
    ViQuAE~\citep{lerner2022viquae}

    \\
    \midrule

    Domains & 
    VQA-Rad~\citep{lau2018dataset}, 
    PMC-VQA~\citep{zhang2024pmcvqavisualinstructiontuning},
    WebSight~\citep{laurençon2024unlocking} ,  \\
    
    &
    CoInstruct~\citep{wu2024openendedvisualqualitycomparison}
    knoiq-10k~\citep{Hosu_2020}

    \\
    \midrule
    Text & 
    Magpie Pro~\citep{Xu2024MagpieAD},
    Magpie Pro (L3 ST),
    Magpie Pro (Qwen2 ST)\\

    &
    AutoMathText~\citep{zhang2024automathtext},
    JiuZhang~\citep{zhou2024jiuzhang30efficientlyimprovingmathematical}
    
    \\
    \bottomrule
    \end{tabular}
    }
    \caption{Datasets/Collections used in stage 3. Some data sources overlap with those in stage 2, and downsample on these sources.}
    \label{tab:data_mixture_stage3}
\end{table*}
\clearpage

\section{Qualitative Examples}
\begin{figure*}[h]
    \centering
    \includegraphics[height=0.85\textheight]{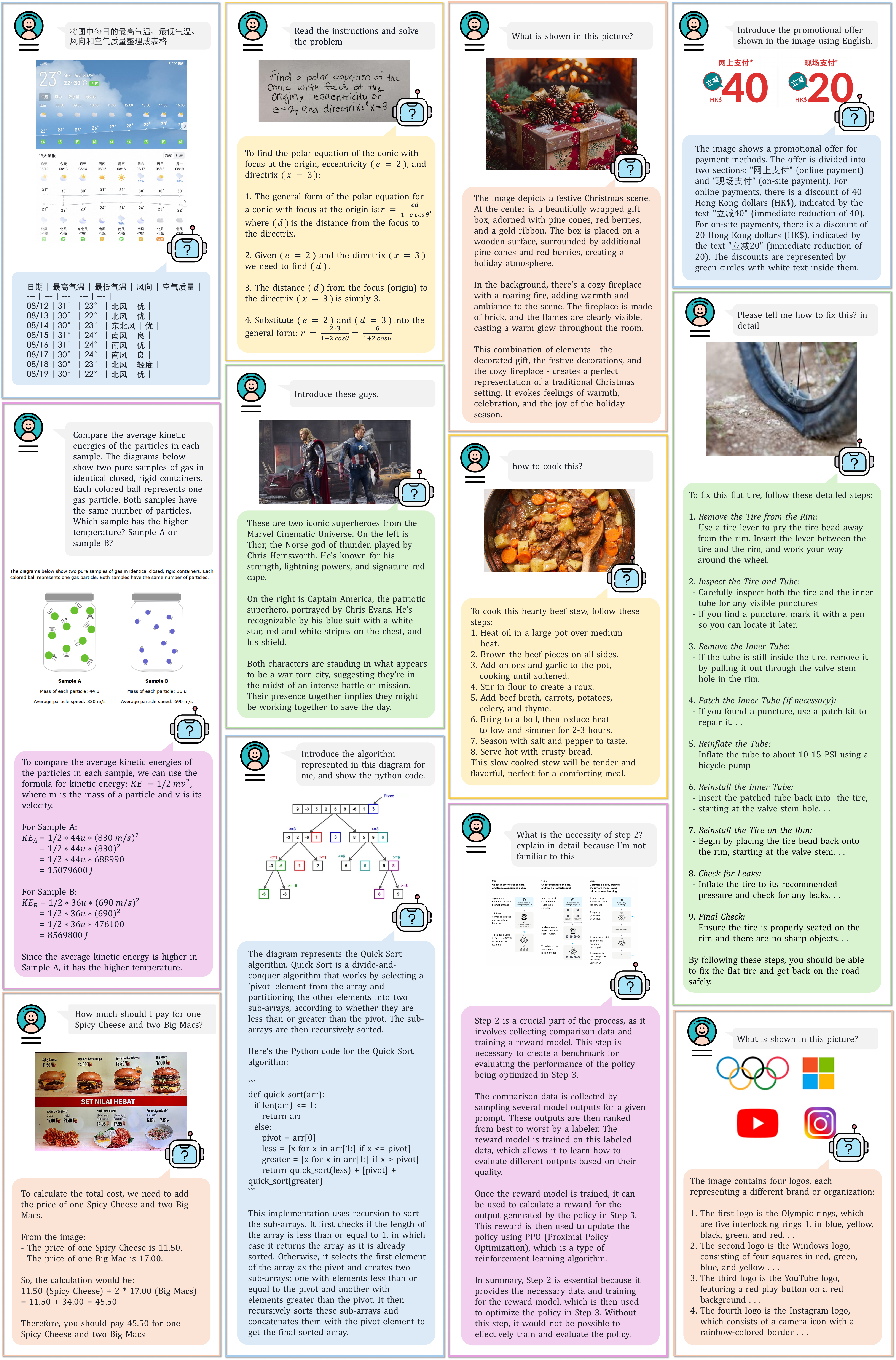}
    \caption{Qualitative examples of \model’s capabilities in image description, information extraction, mathematical and scientific reasoning, multilingual tasks, life assistant functions, etc.}
    \label{fig:goodcase}
\end{figure*}

\end{document}